\documentclass[letterpaper]{article} 
\usepackage{aaai2026}  
\usepackage{times}  
\usepackage{helvet}  
\usepackage{courier}  
\usepackage[hyphens]{url}  
\usepackage{graphicx} 
\usepackage{natbib}  
\usepackage{caption} 
\usepackage{algorithm}
\usepackage{algorithmic}
\usepackage{amssymb}
\usepackage{amsmath}
\usepackage{booktabs}
\usepackage{multirow}

\usepackage{times}
\usepackage{helvet}
\usepackage{courier}
\usepackage{xcolor}
\usepackage{alltt}

\usepackage{newfloat}
\usepackage{listings}
\DeclareCaptionStyle{ruled}{labelfont=normalfont,labelsep=colon,strut=off} 
\floatstyle{ruled}
\newfloat{listing}{tb}{lst}{}
\floatname{listing}{Listing}
\title{A Reasoning Paradigm for Named Entity Recognition}
\author {
	Hui Huang\textsuperscript{\rm 1,\rm2},
	Yanping Chen\textsuperscript{\rm 1,\rm2}\thanks{Corresponding author.},
	Ruizhang Huang\textsuperscript{\rm 1,\rm2},
	Chuan Lin\textsuperscript{\rm 1,\rm2},
	Yongbin Qin\textsuperscript{\rm 1,\rm2}\footnotemark[\value{footnote}]
}
\affiliations {
	\textsuperscript{\rm 1}Text Computing \& Cognitive Intelligence Engineering Research Center of National Education Ministry, College of Computer Science and Technology, Guizhou University, Guiyang, China\\
	\textsuperscript{\rm 2}Big Data Laboratory, Guizhou University, Guiyang, China\\
	\{gs.hhuang21,ypench,rzhuang,clin,ybqin\}@gzu.edu.cn
}

\usepackage{bibentry}

\begin{document}

\maketitle

\begin{abstract}
	Generative LLMs typically improve Named Entity Recognition (NER) performance through instruction tuning. They excel at generating entities by semantic pattern matching but lack an explicit, verifiable reasoning mechanism. This ``cognitive shortcutting'' leads to suboptimal performance and brittle generalization,  especially in zero-shot and low-resource scenarios where reasoning from limited contextual cues is crucial.
	To address this issue, a reasoning framework is proposed for NER, which shifts the extraction paradigm from implicit pattern matching to explicit reasoning. This framework consists of three stages: Chain-of Thought (CoT) generation, CoT tuning, and reasoning enhancement. First, a dataset annotated with NER-oriented CoTs is generated, which contain task-relevant reasoning chains. Then, they are used to tune the NER model to generate coherent rationales before deriving the final answer. Finally, a reasoning enhancement stage is implemented to optimize the reasoning process using a comprehensive reward signal. This stage ensures explicit and verifiable extractions.
	Experiments show that ReasoningNER demonstrates impressive cognitive ability in the NER task, achieving competitive performance. In zero-shot settings, it achieves state-of-the-art (SOTA) performance, outperforming GPT-4 by $12.3$ percentage points on the F1 score. Analytical results also demonstrate its great potential to advance research in reasoning-oriented information extraction.
	Our codes are available at https://github.com/HuiResearch/ReasoningIE.
\end{abstract}


\section{Introduction}

Named entity recognition (NER) is a key task in information extraction, which identifies named entities from unstructured text. Because named entities are phrases that refer to objects of the world,  precisely identifying them is very important for understanding sentence semantics. Traditionally, this task is implemented as a discriminative paradigm, which applies a supervised learning model to classify each token or span to indicate its semantic type. In this paradigm, discriminative language models are widely used to make full use of contextual features and semantic dependencies of a sentence. They usually make a prediction based on statistical models. Therefore,  they are also known as ``conditional models'', which are prone to overfitting the training data and often falter when applied to new domains or unseen entity types \cite{liu2021crossner}.

This limitation has spurred interest in another paradigm for NER: the generative paradigm. In this approach, generative LLMs are employed to leverage their broad knowledge, which is effective to advance the generalization ability of NER models, exhibiting impressive performance in zero-shot settings. This paradigm explores the use of in-context learning to guide models in performing the NER task \cite{wang2023gpt}. However, as the model parameters are not specifically tuned for the NER task, their performance still falls significantly short of specialized models \cite{wang2023instructuie}. To address this gap, researchers focus on instruction-tuning LLMs. For example, InstructUIE \cite{wang2023instructuie} finetunes a T5-11B model on a diverse collection of IE tasks using natural language prompts. A further approach incorporates structured prompts or coding-inspired schemas. For instance, KnowCoder \cite{li2024knowcoder} represents extraction schemas in a code format and uses a two-phase learning process to imbue LLMs with structured knowledge.

\begin{figure}[h]
	\centering
	\includegraphics[scale=0.4]{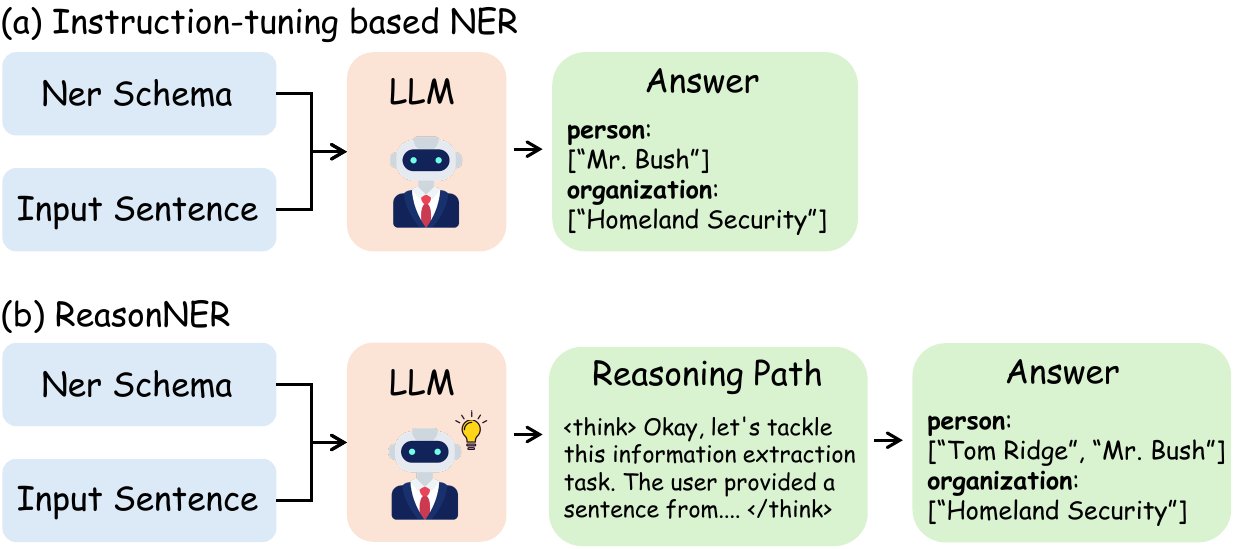}
	\caption{ Comparison between Matching and Reasoning}
	\label{fig:different paradigms}
\end{figure}

Despite this progress, a fundamental limitation persists in both the discriminative and generative paradigms: existing NER models operate primarily through semantic pattern matching rather than explicit reasoning. The difference between the two paradigms is illustrated in Figure \ref{fig:different paradigms}.

As shown in Figure \ref{fig:different paradigms}(a), instruction-tuning prompts an LLM to directly generate entity labels in an opaque, single step. The NER task heavily depends on semantic patterns in the training dataset. The performance often degrades when these models are applied to unseen entities, especially in zero-shot and low-resource scenarios. In contrast, in the reasoning workflow, the extraction process is guided by an explicit, intermediate reasoning path, where a series of reasoning steps are designed to make use of contextual cues to facilitate the NER task. This paradigm incorporates an explicit reasoning mechanism before generating the final answer and can benefit from the vast knowledge and latent reasoning ability of generative LLMs.

Motivated by the above discussions, we propose a reasoning paradigm to support the NER task, named ReasoningNER. This paradigm is composed of three stages: CoT Generation, CoT Tuning, and Reasoning Enhancement. In the first stage, a NER-CoT dataset is constructed, which contains named entities annotated with step-by-step reasoning traces. In the second stage, the NER model is optimized with a CoT analysis, verbalizing how it identifies entities based on contextual clues before outputting the final answer. This  paradigm shifts the task from direct prediction to a reasoned extraction process. In the third stage, a Group Relative Policy Optimization (GRPO) algorithm is employed to further refine the model's policy. Two task-specific reward functions are designed to score the final predicted entities against the ground truth and to verify that the model's output conforms to a predefined schema. The reasoning paradigm enables the extraction process to be guided by contextual clues, semantic hints, and logical constraints, rather than relying purely on semantic patterns.

Extensive experiments are conducted to evaluate the performance and generalization of ReasoningNER. The results show that ReasoningNER demonstrates impressive cognitive ability, attaining competitive performance on multiple NER evaluation datasets. In zero-shot and low-resource settings, ReasoningNER achieves state-of-the-art performance, outperforming GPT-4 by 12.3\% in F1 score. It demonstrates consistent superiority across all target domains and showcases a more robust transfer capability to unseen entity types. Our main contributions are as follows:
\begin{itemize}
	\item A reasoning paradigm is proposed to support NER that formulates the task as a reasoning process. It enables reasoning-oriented information extraction.
	\item A generative NER model (ReasoningNER) is constructed to evaluate the effectiveness of the reasoning paradigm, which shows the potential to support other NLP tasks.
	\item  A NER-CoT corpus is built in this paper in which every entity is accompanied by an explicit chain-of-thought trace. The corpus is helpful to promote research in this field.
\end{itemize}

\section{Related Work}

Recent studies have significantly advanced the capabilities of LLMs on NER by reformulating the task within a generative, prompt-driven paradigm. Similar works like UIE \cite{lu2022unified}, Flan-T5 \cite{chung2024scaling}, and InstructUIE \cite{wang2023instructuie} have demonstrated that multi-task instruction tuning on a diverse set of IE tasks enables models to generalize to novel tasks, achieving performance comparable to BERT-based models in specific scenarios. To further enhance generalization, subsequent research have explored methods for injecting explicit structured knowledge into these models. One line of work focuses on innovative schema representations. For instance, KnowCoder \cite{li2024knowcoder} encodes entity types as Python class definitions, while GoLLIE \cite{sainz2024gollie} directly trains the model to follow human-written annotation guidelines.

Another stream of research concentrates on dynamically providing contextual information at inference time rather than relying on the model's static, internalized knowledge. RUIE \cite{liao2025ruie} employs a retrieval-augmented mechanism to furnish input samples with relevant demonstrations, whereas GPT-NER \cite{wang2023gpt} leverages few-shot prompting and a self-verification step to guide the model. Although these approaches have achieved considerable success in improving the flexibility of NER, their mechanism remains a direct mapping from instructions to labels. They lack an explicit reasoning process, which limits their reliability in complex and ambiguous scenarios that demand deep contextual understanding or logical inference.

CoT reasoning, which prompts LLMs to generate a series of intermediate steps to solve complex problems, unlocks the potential of LLMs in tasks such as arithmetic and commonsense reasoning \cite{wei2022chain,wang2022self,ma2023chain,agarwal2025think}. The central idea is to transform the implicit, single-step ``black-box'' prediction into an explicit, decomposable reasoning trajectory, thereby enhancing model accuracy and interpretability.
However, the application of CoT to IE remains nascent. A few pioneering studies have begun to integrate reasoning into IE. For example, PromptNER \cite{ashok2023promptner} decomposes NER into ``judgment'' and ``extraction'' steps, offering initial validation for the efficacy of simple reasoning prompts. Similarly, ERA-CoT \cite{liu2024era} addresses multi-entity relation problems by first extracting all entities and then applying CoT to reason over this structured information.

These studies suggest that guiding a model to ``think'' about its extraction decisions is beneficial. However, current models primarily depend on prompts at inference time. They fail to supervise the generation of a reasoning chain as a core part of the model's functionality. In the information extraction field, there is a lack of a systematic framework for training models to generate and utilize a verifiable reasoning process. By explicitly supervising the reasoning steps, our proposed method combines the advantages of CoT with the generalization power of instruction tuning to achieve a more robust and transparent reasoning-driven approach.

\begin{figure*}[t]
	\centering
	\includegraphics[scale=0.59]{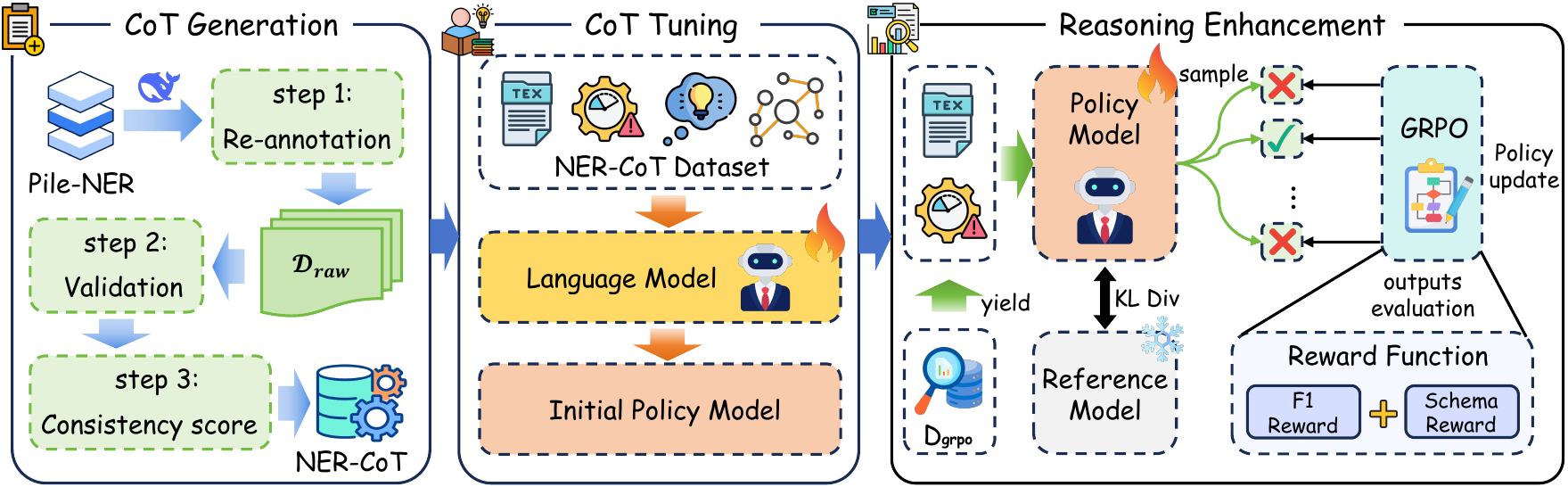}
	\caption{The architecture of the ReasoningNER model}
	\label{fig:framework}
\end{figure*}

\section{Methodology}

The architecture of the ReasoningNER model is composed of three stages: CoT Generation (CG), CoT Tuning (CT) and Reasoning Enhancement (RE). The architecture is illustrated in Figure \ref{fig:framework}. The CG stage constructs a semi-supervised NER-CoT dataset for training the model. In the CT stage, a supervised fine-tuning is applied to establish the model reasoning capability.The RE stage enhances the reasoning ability through a reinforcement learning. In the sections that follow, we will first provide a problem formulation, then describe each stage in details.

\subsection{Problem Formulation}

Let $X = [w_1, w_2, \dots, w_n]$ be an input text composed of $n$ tokens, $\mathcal{S} = \{\ell_1, \ell_2, \dots, \ell_m\}$ is an entity schema, where $\ell_i$ denotes to a predefined entity type.

All named entities in $X$ can be refer as a set:
\begin{equation}
E = \{ (s_i, \ell_j) | s_i\subset X, \ell_j \in  \mathcal{S}\},
\end{equation}
where $s_i$ is a contiguous subsequence $[w_p, \dots, w_q] $ ($1 \leq p \leq q \leq n$), $\ell_j \in \mathcal{S}$ is a entity type of $s_i$.

A CoT $\mathcal{C}=[c_1,c_2,\dots,c_L]$ is a depiction composed of $L$ tokens, that provides the reasoning path for identifying and classifying entities in $X$.

The objective of reasoning NER is to produce an output $Y=(\mathcal{C},E)$ that encompasses a CoT and entity set belonging to $X$.

\subsection{CoT Generation}

Before generating the NER-CoT dataset, a NER dataset should be adopted to produce not only extracted entities but also the explicit reasoning path for each sentence. In our work, the Pile-NER corpus \cite{zhou2023universalner} is used, which contains a collection of sentences annotated with fine-grained entity schemas.

Based on this dataset, as illustrated in the CG stage on the left of Figure \ref{fig:framework}, the construction contains three steps: {\em Re-Annotation}, {\em Validation} and {\em Consistency}.

Re-annotation Step: Based on sentences and entity schemas in Pile-NER, a specialized prompt template is designed to guide a LLM (DeepSeek-R1 \cite{guo2025deepseek}) to produce not only entities but also explicit reasoning paths. An example of this prompt template is provided in Appendix. This step yields an initial, raw collection of CoT annotations, denoted as $\mathcal{D}_{\mathrm{raw}}$. Each instance in $\mathcal{D}_{\mathrm{raw}}$ is a tuple $(X, \mathcal{S}, \mathcal{C}_{\mathrm{raw}}, E_{\mathrm{raw}})$, where $\mathcal{C}_{\mathrm{raw}}$ is the raw reasoning chain and $E_{\mathrm{raw}}$ is the list of extracted entities.

Validation Step: because the outputs $\mathcal{D}_{\mathrm{raw}}$ may contain logical fallacies or structural errors, we propose a strict structural integrity to validate the initial outputs. In this validation step, a sample is retained only if its reasoning trace $\mathcal{C}$ explicitly justifies every entity in the list $E$, and the list $E$ is well-formatted and compliant with the predefined entity schema $\mathcal{S}$. Samples failing these consistency checks are discarded.

Consistency Step: in this step, we move beyond structural checks to evaluate the semantic quality of the reasoning path. An auxiliary LLM (Qwen3 32B) is employed to assess the internal coherence, logical soundness, and factual accuracy of the reasoning process $\mathcal{C}$ relevant to the input $X$ and the extracted entities $E$. This evaluation yields a quantitative consistency score on a scale of 0 to 10. By setting a high threshold of 9, we ensure that only the most coherent and logically sound samples are kept.

This CoT generation process builds the final NER-CoT dataset, denoted as $\mathcal{D}_{\mathrm{cot}}$, which is used for the subsequent CoT Tuning and Reasoning Enhancement stages.

\subsection{CoT Tuning}

In this stage, the NER-CoT dataset is used to increase the reasoning capability of a LLM.

Each sample in the NER-CoT dataset is a quadruple $(X, \mathcal{S}, \mathcal{C}, E)$, where $X$ is the input sentence, $\mathcal{S}$ is the entity schema, $\mathcal{C}$ is the CoT reasoning trace, and $E$ is the final list of entities. The output of the model is  $Y=(\mathcal{C},E)$. It can be represented as a token sequence: $Y = \mathcal{C} \oplus E = [y_1, y_2, \dots, y_T]$ formed by concatenating the reasoning trace $\mathcal{C}$ and the entity list $E$.

The training objective is to maximize the conditional likelihood of this target sequence given the input sentence $X$ and schema $\mathcal{S}$. This is achieved by minimizing the negative log-likelihood (NLL) loss with respect to the model parameters $\theta$:
\begin{equation}
\mathcal{L}_{\mathrm{SFT}}(\theta) = - \mathbb{E}_{(X, \mathcal{S}, Y) \sim \mathcal{D}_{\mathrm{CoT}}} \left[ \sum_{t=1}^{T} \log \pi_{\theta}(y_t | X, \mathcal{S}, y_{<t}) \right],
\end{equation}
where $\pi_{\theta}$ denotes the language model with parameters $\theta$, and $y_t$ is the $t$-th token in the target sequence $Y$. This CoT Tuning process yields the Initial Policy Model, referred as $\pi_{\theta}$, as shown in Figure \ref{fig:framework}. 

\subsection{Reasoning Enhancement}

In this stage, the Initial policy model ($\pi_{\theta}$) obtained from CoT Tuning stage is tuned by a GRPO algorithm \cite{guo2025deepseek}. To prevent the model from deviating too far from the learned CoT behavior, a reference model ($\pi_{\mathrm{ref}}$) is built to regularize the policy updates. We initialize both the policy model, which is the policy to be optimized, and the reference model with the weights of the initial model. 

A dataset for GRPO training, denoted $\mathcal{D}_{\mathrm{grpo}}$, is constructed by stratified sampling without replacement from the InstructUIE dataset. It generates a fixed-size subset of 4,703 samples. For each query $q=(X,\mathcal{S})$ in $\mathcal{D}_{\mathrm{grpo}}$, a group of $G$ candidate outputs $\{o_1, o_2, \dots, o_G\}$ is sampled from the old policy $\pi_{\theta_{old}}$. Each output $o_i$ comprises a reasoning chain $\mathcal{C}_i$ and a final list of entities $E_i$, i.e., $o_i=(\mathcal{C}_i,E_i)$. Subsequently, each output is evaluated based on a reward function. 

As shown in Figure \ref{fig:framework}, generated outputs are sent to the ``outputs evaluation'' step and scored by a Reward Function. To guide the model towards optimal NER performance, a composite reward is designed to assess both task accuracy and schema adherence. First, a span-level F1 reward, $R_{F1}$, quantifies the match between the extracted entities and the ground truth. It is computed as the micro-averaged F1 score, linearly scaled to the range $[0,1]$, providing a direct incentive for precision and recall. Second, a structural consistency reward, $R_{schema}$, evaluates adherence to the predefined output format and entity type constraints. Outputs with any schema violations (e.g., malformed structures or unpermitted entity types) receive $R_{schema} = 0$, whereas compliant outputs receive $R_{schema} = 1$.

The total reward $R(o_i)$ is a weighted combination:
\begin{equation}
R(o_i) = {\lambda}_{F1} R_{F1}(o_i) + {\lambda}_{schema}R_{schema}(o_i).
\end{equation}
In our experiments, the coefficients ${\lambda}_{F1}$ and ${\lambda}_{schema}$ are set to 10 and 1, respectively, to prioritize entity correctness while strongly enforcing structural validity.

Finally, the computed rewards guide the GRPO Policy update. The average reward of the samples within the group is utilized as a baseline to compute the relative advantage. Let $\bar{R}$ be a mean reward within the group, computed as:
\begin{equation}
\bar{R} = \frac{1}{G}\sum_{j=1}^{G} R(o_j).
\end{equation}

Then, the advantage $A_i$ for $o_i$ is defined as:
\begin{equation}
A_i=R(o_i)-\bar{R}.
\end{equation}

This advantage $A_i$ is used for policy updates. The policy parameters $\theta$ are optimized by maximizing the following objective function:
\begin{align}
	\mathcal{J}_{\mathrm{GRPO}}(\theta) &= \mathbb{E}_{q,\{o_{i}\}\sim\pi_{\theta_{\mathrm{old}}}} \left[ \frac{1}{G} \sum_{i=1}^G \left(\mathcal{L}_i-\beta\mathbb{D}_{\mathrm{KL}}\left(\pi_\theta||\pi_{\mathrm{ref}}\right)\right) \right]\\
	\mathcal{L}_i &= \min\left(r_iA_i,\mathrm{clip}\left(r_i,1-\varepsilon,1+\varepsilon\right)A_i\right)\\
	r_i &= \frac{\pi_\theta(o_i|q)}{\pi_{\theta_{old}}(o_i|q)},
\end{align}
where $\pi_{\theta_{old}}$ is the old policy, $\pi_{\theta}$ is the current policy being optimized, and $\varepsilon$ is a clipping hyperparameter. The $\mathbb{D}_{KL}$ term represents KL divergence regularization, penalizing deviations of the current policy $\pi_{\theta}$ from a reference policy $\pi_{ref}$. 

\section{Experiments}

\subsection{Datasets}

Our framework utilizes data across two training stages. First, for the CoT Tuning stage, we use our proposed NER-CoT dataset, which contains 45,787 training samples. Second, for the Reasoning Enhancement stage, we constructed a 4,703-sample dataset by applying stratified sampling to the InstructUIE \cite{wang2023instructuie}. This collection aggregates 20 diverse NER datasets. To ensure a balanced representation and prevent large-scale datasets from dominating the sampling proportions, we imposed a cap, limiting the size of any dataset to 10,000 when calculating its sampling ratio. The 4,703 samples were then drawn proportionally from all 20 datasets based on these adjusted, capped sizes.

To evaluate model performance, we use a suite of seven datasets. To assess out-of-domain generalization, we employ five datasets from the Cross-NER \cite{liu2021crossner}: artificial intelligence, literature, music, politics, and science. To evaluate in-domain performance, we use the MIT-Movie and MIT-Restaurant datasets \cite{liu2021crossner}. These two datasets are considered in-domain because their training data is part of the InstructUIE collection used in our RE sampling pool. Consistent with previous research \cite{wadden2019entity,lin2020joint,van2021cross,lu2022unified,li2024knowcoder}, we use span-based Micro-F1 as the evaluation metric.

\subsection{Implementation Details}

For our experiments, the qwen3-8B-Base \cite{qwen3} model is optimized with AdamW. The initial supervised fine-tuning is conducted for five epochs with a sequence length of 8192, a learning rate of $2 \times 10^{-5}$ using a cosine scheduler, and a batch size of 256. Subsequently, we performed reinforcement learning using GRPO for one epoch, with a clipping threshold $\epsilon$ of 0.2 and a KL coefficient $\beta$ of 0.04, maintaining an effective batch size of 384. To enhance computational efficiency, we integrated bfloat16 mixed-precision, gradient checkpointing, FlashAttention-2\cite{dao2023flashattention}, and the Liger-kernel\cite{hsu2024liger}. For detailed hyper-parameter settings, please refer to Appendix.

\subsection{Main Results}

\begin{table*}[t]
	\centering
	\caption{Out-of-domain evaluation results. ``\#Num'' denotes  the number of Information Extraction related samples used for model training, and ``Avg'' denotes the average performance across the seven datasets. $\clubsuit$ denotes the results evaluated on the test set processed by $\text{B}^2\text{NER}$\cite{yang2025beyond}.}
	\label{table:main-exp}
	\scalebox{0.94}{
		\begin{tabular}{l|c|ccccccc|c}
			\toprule
			\textbf{Model}            & \textbf{\#Num} & \textbf{Movie.} & \textbf{Rest.} & \textbf{AI}   & \textbf{Litera.} & \textbf{Music} & \textbf{Politics} & \textbf{Science} & \textbf{Avg}  \\ \midrule
			\multicolumn{10}{l}{\textit{\textbf{Non Reasoning Model}}}                                                                                                                                            \\
			ChatGPT-3.5\cite{ouyang2022training}                   & -                  & 5.3             & 32.8           & 52.4          & 39.8             & 66.6           & 68.5              & 67.0             & 47.5          \\
			UniNER 7B\cite{zhou2023universalner}                 & 45,884             & 42.4            & 31.7           & 53.5          & 59.4             & 65.0           & 60.8              & 61.1             & 53.4          \\
			UniNER 13B\cite{zhou2023universalner}                & 45,884             & 48.7            & 36.2           & 54.2          & 60.9             & 64.5           & 61.4              & 63.5             & 55.6          \\
			InstructUIE 11B\cite{wang2023instructuie}           & 225,930            & -               & -              & 48.4          & 48.8             & 54.4           & 49.9              & 49.4             & -             \\
			GoLLIE 7B\cite{sainz2024gollie}                 & 165,163            & 63.0            & 43.4           & 59.1          & 62.7             & 67.8           & 57.2              & 55.5             & 58.4          \\
			GoLLIE 13B\cite{sainz2024gollie}                & 165,163            & 62.5            & 49.8           & 56.7          & 59.7             & 65.5           & 54.4              & 56.2             & 57.8          \\
			KnowCoder 7B\cite{li2024knowcoder}             & 4,590,403          & 50.0            & 48.2           & 60.3          & 61.1             & 70.0           & 72.2              & 59.1             & 60.1          \\
			GPT-4\cite{achiam2023gpt}                     & -                  & 60.4            & 59.7           & 50.0          & 55.2             & 69.2           & 63.4              & 63.2             & 60.1          \\
			GLiNER-L\cite{zaratiana2024gliner}                  & 44,889             & 57.2            & 42.9           & 57.2          & 64.4             & 69.6           & 72.6              & 62.6             & 60.9          \\
			$\text{B}^2\text{NER 7B}^\clubsuit$\cite{yang2025beyond} & 70,292             & 67.6            & 53.3           & 59.0          & 63.7             & 68.6           & 67.8              & 72.0             & 64.6          \\ \midrule
			\multicolumn{10}{l}{\textit{\textbf{Reasoning Model}}}                                                                                                                                                \\
			deepseek-r1 8B\cite{guo2025deepseek}            & -                  & 60.1            & 50.6           & 47.6          & 57.4             & 53.3           & 58.1              & 55.3             & 54.6          \\
			deepseek-r1 32B\cite{guo2025deepseek}           & -                  & 70.4            & \textbf{57.5}  & 60.4          & 52.3             & 70.4           & 71.1              & 65.9             & 64.0          \\
			qwen3 4B\cite{qwen3}                  & -                  & 69.6            & 53.4           & 59.2          & 56.8             & 62.3           & 68.4              & 65.4             & 62.1          \\
			qwen3 8B\cite{qwen3}                  & -                  & 70.1            & 57.4           & 61.2          & 58.0             & 71.0           & 71.9              & 68.6             & 65.4          \\ \midrule
			ReasoningNER 1.7B(Ours)        & 50,787             & 70.2            & 52.4           & 63.6          & 59.1             & 71.6           & 68.6              & 69.9             & 65.1          \\
			ReasoningNER 8B(Ours)           & 50,787             & \textbf{76.3}   & 56.8           & \textbf{71.0} & \textbf{69.4}    & \textbf{78.7}  & \textbf{78.8}     & \textbf{75.8}    & \textbf{72.4} \\
			$\text{ReasoningNER 8B}^\clubsuit$(Ours)           & 50,787             & \textbf{79.3}   & \textbf{67.7}           & \textbf{72.2} & \textbf{77.1}    & \textbf{84.0}  & \textbf{79.8}     & \textbf{81.4}    & \textbf{77.3} \\ \bottomrule
		\end{tabular}
	}
\end{table*}

The efficacy of ReasoningNER is substantiated by its performance on the CrossNER benchmark, as detailed in Table \ref{table:main-exp}. In the zero-shot cross-domain setting, ReasoningNER achieves new state-of-the-art results, with ReasoningNER 8B attaining an average F1 score of 72.4 across all seven domains. Notably, even our smaller ReasoningNER 1.7B model demonstrates competitive performance, achieving an average F1 of 65.1 and surpassing many larger models.

\textbf{Comparison with Non-Reasoning models.} In the cross-domain setting, ReasoningNER significantly outperforms all non-reasoning baselines. As shown in Table \ref{table:main-exp}, ReasoningNER 8B achieves a new state-of-the-art average F1 score of 72.4 on the CrossNER benchmark, surpassing $\text{B}^2\text{NER}$ 7B by 12.7 points and the larger GPT-4 by 12.3 points. This highlights a fundamental limitation of conventional models. The origin of this issue lies in the instruction-tuning paradigm, where models are taught a direct mapping from input to final entity labels, omitting explicit analytical steps. This training objective incentivizes the model to find the shortest path to the answer, a behavior we identify as ``cognitive shortcutting.'' While efficient at recognizing patterns in training data, this shortcut is the primary cause of their brittle generalization, leading to precipitous performance degradation when faced with misleading surface-level patterns in unseen domains. In contrast, ReasoningNER leverages explicit reasoning guided by CoT to comprehend contextual logic, enabling it to accurately identify entities in novel domains and effectively overcome the limitations of methods that depend on direct input-output mappings.

\textbf{Comparison with Reasoning Models.} ReasoningNER 8B also demonstrates a clear advantage over models with innate reasoning capabilities, outperforming Qwen3 8B by 7.0 average F1 points. This validates the efficacy of our proposed methods, which imbue the base model with NER-specific reasoning abilities that are more potent than its native, general-purpose architecture. Moreover, ReasoningNER 8B surpasses the much larger DeepSeek-R1 32B by 8.4 F1 points, while our smaller 1.7B model substantially outperforms DeepSeek-R1 8B by 10.5 points. This proves our two-stage targeted optimization (SFT and GRPO) creates a critical task-specific alignment, enabling ReasoningNER to achieve superior performance over models that possess general reasoning abilities but lack focused optimization for resolving ambiguity in complex NER contexts.

\subsection{Ablation Study}

In this section, we analyze the impact of the NER-CoT dataset, the introduction of CoT during SFT, and the subsequent RE phase. As shown in Table \ref{table:ablation study}, our analysis yields the following observations: (1) SFT on the Pile-NER dataset boosts the F1 score of the Qwen3-1.7B-Base model to 40.3, an increase of 8.4 points. This improvement underscores the foundational importance of supervised fine-tuning on task-specific data. (2) Notably, training solely on our constructed NER-CoT annotations, even without CoT reasoning, achieves a substantial 20.3-point F1 improvement over training on the standard Pile-NER dataset. This result highlights the superior quality of our reconstructed annotations. (3) Incorporating CoT reasoning during the SFT process further increases the F1 score by 2.4 points, confirming that explicitly modeling the reasoning process enables the model to capture more complex entity patterns. (4) Finally, the RE phase provides an additional 2.1-point F1 gain, demonstrating that fine-tuning the reasoning pathways with a task-aligned reward signal via GRPO effectively refines the model's predictions and improves overall performance.

\begin{table}[h]
	\centering
	\caption{Ablation study showing the average F1 score across the CrossNER and MIT datasets. We start with the Qwen3-1.7B-Base model and incrementally add: (1) SFT on Pile-NER; (2) SFT on our NER-CoT dataset annotations (w/o CoT); (3) SFT with CoT reasoning; and (4) the final Reasoning Enhancement (RE) phase.}
	\label{table:ablation study}
	\begin{tabular}{l|c}
		\toprule
		\textbf{Model}        & \textbf{F1}         \\ \midrule
		Qwen3-1.7B-Base                & 31.9                \\ 
		+ SFT on Pile-NER                &       40.3               \\
		+ SFT on NER-CoT (w/o cot)      & 60.6 \\
		+ SFT on NER-CoT (w/ cot)       & 63.0                 \\
		+ SFT on NER-CoT (w/ cot) \& RE & 65.1                 \\ \bottomrule
	\end{tabular}
	
\end{table}

\subsection{Analysis}

\subsection{Analysis of Supervised Evaluation}

For in-domain supervised evaluation, we address the absence of CoT annotations across the 20 NER benchmarks by creating a mixed-inference training set \cite{qwen3}. We first construct a targeted subset by sampling 50 examples per category, along with 50 negative examples, from each of the 20 datasets. On this curated subset, we apply GRPO reinforcement learning for 3 epochs to train a specialized policy model capable of generating reasoning paths. This model then performs inference on the full training set to generate synthetic CoT. Through a filtering process, we retain these reasoning paths only for samples where the model's predictions are verifiably correct, yielding a high-fidelity, CoT-augmented dataset. This is then merged with the remaining non-CoT samples to form a hybrid corpus, enabling our final ReasoningNER model to be fine-tuned with a combination of explicit reasoning and standard supervision.

The results of this strategy, presented in Table \ref{table:supervised}, demonstrate its efficacy. Our ReasoningNER model achieves a new state-of-the-art average F1 score of 85.2, outperforming strong baselines like $\text{B}^2\text{NER}$ by 1.3 percentage points and securing top performance on 11 of the 20 datasets. Remarkably, this significant improvement is achieved even though the CoT-augmented data constitutes only a small fraction of the training corpus. This indicates that even a limited exposure to explicit reasoning provides a powerful inductive bias, enhancing the model's learning dynamics and sample efficiency, while the larger set of standard examples ensures robust generalization. We plan to explore methods for creating more comprehensive CoT data in future work to unlock greater performance.

\begin{table}[h]
	\small
	\centering
	\caption{Results of in-domain supervised evaluation on 20 NER benchmarks.}
	\label{table:supervised}
	\setlength{\tabcolsep}{1mm}
		\begin{tabular}{c|cccc|c}
			\toprule
			Dataset          & InstructUIE & UniNER & GLiNER & $\text{B}^2\text{NER}$ & Ours \\ \midrule
			ACE05            & 79.9                 & \textbf{86.7}   & 82.8              & 83.0                                & 86.6      \\
			AnatEM           & 88.5                 & 88.7            & 88.9              & 89.2                                & \textbf{89.9}      \\
			bc2gm            & 80.7                 & 82.4            & 83.7              & 82.0                                & \textbf{84.5}      \\
			bc4chemd         & 87.6                 & 89.2            & 87.9              & 89.0                                & \textbf{91.6}      \\
			bc5cdr           & 89.0                 & 89.3            & 88.7              & 88.5                                & \textbf{90.1}      \\
			Broad Twitter    & 80.3                 & 81.3            & \textbf{82.5}     & 82.2                                & 81.2      \\
			CoNLL03          & 91.5                 & 93.3            & 92.6              & 92.6                                & \textbf{94.4}      \\
			FabNER           & 78.4                 & 81.9            & 77.8              & 78.8                                & \textbf{82.4}      \\
			FindVehicle      & 87.6                 & 98.3            & 95.7              & 97.9                                & \textbf{98.4}      \\
			GENIA            & 75.7                 & 77.5            & \textbf{78.9}     & 76.4                                & 78.3      \\
			HarveyNER        & \textbf{74.7}        & 74.2            & 68.6              & 73.7                                & 74.2      \\
			Movie.        & 89.6                 & 90.2            & 87.9              & \textbf{90.8}                       & 89.6      \\
			Restaurant.   & 82.6                 & 82.4            & 83.6              & \textbf{83.7}                       & 81.4      \\
			MultiNERD        & 90.3                 & 93.7            & 93.8              & 94.0                                & \textbf{94.5}      \\
			ncbi             & 86.2                 & 87.0            & 87.8              & 84.8                                & \textbf{87.9}      \\
			OntoNotes        & 88.6                 & \textbf{89.9}   & 89.0              & 84.3                                & 89.1      \\
			PolyglotNER      & 53.3                 & 65.7            & 61.5              & 62.0                                & \textbf{65.8}      \\
			TweetNER7        & \textbf{66.0}        & 65.8            & 54.4              & 66.3                                & 65.8     \\
			WikiANN          & 64.5                 & 84.9            & 83.7              & 85.1                                & \textbf{85.9}      \\
			wikiNeural       & 88.3                 & \textbf{93.3}   & 91.3              & 93.0                                & 92.8      \\ \midrule
			\textbf{Average} & 81.2                 & 84.8            & 82.9              & 83.9                                & \textbf{85.2}      \\ \bottomrule
		\end{tabular}
\end{table}

\subsubsection{Analysis of Zero-Shot Performance}

We conducted a zero-shot performance evaluation on these same 20 datasets.  The results, as shown in Table \ref{table:zero-shot}, are directly from the ReasoningNER 8B model that has only undergone the CT stage. This model was not trained on any of these 20 datasets. ReasoningNER establishes a new SoTA in zero-shot NER, achieving an average F1 score of 56.8 across 20 benchmark datasets. This performance represents a significant leap over existing methods, outperforming the next-best model, GLiNER-L, by 9 F1 points and demonstrating a 11.1-point improvement over UniNER. The model's consistently superior performance across all 20 diverse datasets, without any task-specific training, substantiates the efficacy of our proposed reasoning-centric learning paradigm. By first internalizing explicit reasoning steps through supervised fine-tuning on the NER-CoT corpus and subsequently refining these reasoning pathways via GRPO, ReasoningNER acquires a generalizable, procedural approach to entity identification. This methodology moves beyond superficial pattern matching, enabling the model to develop a more robust and adaptable understanding of entity concepts, which translates into its zero-shot generalization capabilities.

\begin{table}[h]
	\centering
	\caption{Zero-shot performance of ReasoningNER 8B compared against other models on 20 NER benchmarks. All results are F1 scores, and the best results are bolded. Results of ChatGPT are reported from UniNER \cite{zhou2023universalner}.}
	\label{table:zero-shot}
	\setlength{\tabcolsep}{1mm}
		\begin{tabular}{c|ccc|c}
			\toprule
			Dataset          & ChatGPT & UniNER & GLiNER-L & \textbf{Ours} \\ \midrule
			ACE05            & 26.6    & \textbf{36.9}   & 27.3     & 33.8     \\
			AnatEM           & 30.7    & 25.1   & 33.3     & \textbf{49.5}      \\
			bc2gm            & 40.2    & 46.2   & \textbf{47.9}     & 43.3     \\
			bc4chemd         & 35.5    & 47.9   & 43.1     & \textbf{53.1}      \\
			bc5cdr           & 52.4    & 68.0   & 66.4     & \textbf{74.8}      \\
			Broad Twitter    & 61.8    & 67.9   & 61.2     & \textbf{76.0}      \\
			CoNLL03          & 52.5    & 72.2   & 64.6     & \textbf{79.0}      \\
			FabNER           & 15.3    & 24.8   & 23.6     & \textbf{26.0}      \\
			FindVehicle      & 10.5    & 22.2   & 41.9     & \textbf{57.4}      \\
			GENIA            & 41.6    & 54.1   & 55.5     & \textbf{58.9}      \\
			HarveyNER        & 11.6    & 18.2   & 22.7     & \textbf{26.8}      \\
			MIT Movie        & 5.3     & 42.4   & 57.2     & \textbf{76.0}      \\
			MIT Restaurant   & 32.8    & 31.7   & 42.9     & \textbf{57.6}      \\
			MultiNERD        & 58.1    & 59.3   & 59.7     & \textbf{65.5}      \\
			ncbi             & 42.1    & 60.4   & 61.9     & \textbf{69.9}      \\
			OntoNotes        & 29.7    & 27.8   & 32.2     & \textbf{44.2}      \\
			PolyglotNER      & 33.6    & 41.8   & 42.9     & \textbf{49.4}      \\
			TweetNER7        & 40.1    & 42.7   & 41.4     & \textbf{52.5}      \\
			WikiANN          & 52.0    & 55.4   & \textbf{58.9}     & 58.6      \\
			wikiNeural       & 57.7    & 69.2   & 71.8     & \textbf{83.4}      \\ \midrule
			\textbf{Average} & 36.5    & 45.7   & 47.8   & \textbf{56.8}      \\ \bottomrule
		\end{tabular}
\end{table}

\subsubsection{Analysis of Few-Shot Performance}

\begin{figure}[t]
	\centering
	\includegraphics{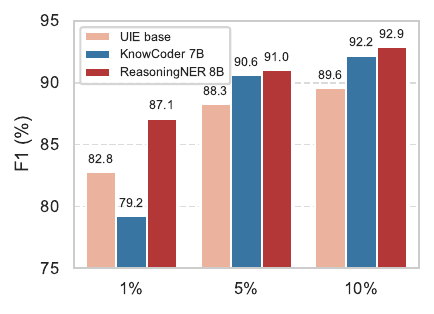}
	\caption{Low-resource results on CoNLL03 dataset.}
	\label{fig:few-shot}
\end{figure}

To evaluate the model's sample efficiency in low-resource settings, we followed the experimental setup of UIE: the CoNLL03 training set was randomly downsampled to 1\%, 5\%, and 10\% of its original size, and these data subsets were used to continue training the ReasoningNER 8B model. We compare it against two baseline models, UIE-base and KnowCoder-7B, with the results shown in Figure \ref{fig:few-shot}.
Across all data proportions, the performance of ReasoningNER 8B surpasses both baseline models. This performance advantage is most significant in the most data-scarce setting (1\%), where our model achieves an F1 score of 87.1\%, outperforming UIE-base by 4.3 points and KnowCoder-7B by  7.9 points. When the amount of training data increases to 5\% and 10\%, ReasoningNER 8B maintains the best performance, achieving F1 scores of 91.0\% and 92.9\%, respectively. These results indicate that while large-scale instruction tuning contribute to improving model generalization, the fine-grained logical reasoning supervision employed in our model is crucial for maximizing sample efficiency in low-resource scenarios.

\subsubsection{Analysis of the Impact of Backbone LLMs}
To ascertain that the superior performance of ReasoningNER is not contingent upon a specific LLM, we conducted a comprehensive evaluation across a suite of mainstream models with varying architectures and scales. For this analysis, all models were uniformly fine-tuned exclusively during the SFT phase, thereby circumventing the computationally intensive GRPO process. 
As detailed in Table \ref{table:ana-backbone}, when trained on a modest 45,787 samples from our NER-CoT dataset, all ReasonNER variants demonstrate a significant performance advantage over three competitive baselines. Employing the same InternLM2-7B model and test set, our method achieves an F1 score of 74.9, surpassing $\text{B}^2\text{NER}$'s 64.6. Similarly, the Llama2-7B variant reaches 63.5 F1, surpassing KnowCoder (60.1 F1) despite the latter being trained on a dataset two orders of magnitude larger. Furthermore, we observe a monotonic improvement that correlates with model capacity: the Qwen3 series scales from 63.0 F1 (1.7B) to 69.0 F1 (4B), ultimately achieving a peak performance of 70.2 F1 with the 8B model. This result surpasses both InternLM2-7B (69.1 F1) and Llama3.1-8B (68.9 F1).
These findings suggest that incorporating explicit chains of reasoning is a effective strategy for unlocking the full potential of various LLMs and demonstrates substantially greater data efficiency than large-scale instruction tuning. Given its superior performance in this comparative analysis, we have selected Qwen3-8B as the backbone for our main experiments, where it is further optimized with GRPO to establish a new SoTA.

\begin{table}[t]
	\centering
	\caption{Performance Comparison of Supervised Fine-Tuning on the NER-CoT Dataset across Various LLMs, Measured by the Average F1-Score on Seven Datasets from CrossNER and MIT. $\clubsuit$ denotes the results evaluated on the test set processed by $\text{B}^2\text{NER}$\cite{yang2025beyond}}
	\label{table:ana-backbone}
	\begin{tabular}{l|c|c}
		\toprule
		\textbf{Model} & \textbf{\#Num} & \textbf{F1} \\
		\midrule
		\multicolumn{3}{l}{\textit{\textbf{SFT on NER-CoT}}} \\
		Qwen3-8B & 45,787 & \textbf{70.2} \\
		Qwen3-8B$^\clubsuit$ & 45,787 & \textbf{75.9} \\
		Qwen3-4B        & 45,787 & 69.0 \\
		Qwen3-1.7B      & 45,787 & 63.0 \\
		InternLM2-7B & 45,787 & 69.1 \\
		InternLM2-7B$^\clubsuit$ & 45,787 & 74.9 \\
		Llama3.1-8B     & 45,787 & 68.9 \\
		Qwen2.5-7B      & 45,787 & 68.6 \\
		Llama2-7B       & 45,787 & 63.5 \\
		\midrule
		\multicolumn{3}{l}{\textit{\textbf{Baselines}}} \\
		GPT-4 (Zero-shot) & -     & 60.1 \\
		KnowCoder (Llama2-7B) & 4,590,403 & 60.1 \\
		$\text{B}^2\text{NER}^\clubsuit$ (InternLM2-7B) & 70,292  & 64.6 \\
		\bottomrule
	\end{tabular}
	
\end{table}

\section{Conclusion}

This paper introduces ReasoningNER, an NER framework grounded in a reasoning paradigm. It reframes the task from implicit pattern matching to explicit, step-by-step logical inference. This paradigmatic shift significantly enhances the generalization and data efficiency of large language models. To operationalize this paradigm, we present a three-stage approach. First, we construct NER-CoT, the first corpus for NER where entities are annotated with fine-grained Chain-of-Thought rationales. Second, we employ CoT Tuning to supervise the model in generating structured reasoning paths before making a prediction. Finally, a Reasoning Enhancement stage utilizes the GRPO algorithm to directly optimize the reasoning policy, guided by a composite reward function that ensures both task accuracy and structural-format consistency. Empirically, ReasoningNER achieves state-of-the-art results, significantly outperforming existing methods across various benchmarks and demonstrating superior robustness in zero-shot, cross-domain, and low-resource scenarios. Despite its validated efficacy, a primary limitation is the increased inference latency from generating longer reasoning chains. Future work will explore hybrid CoT strategies to strike a better balance between model performance and inference efficiency.

\bibliography{aaai2026}

\newpage

\appendix

\section{Data Details}
\label{appendix-data}

This section provides a detailed overview of the datasets utilized in the various stages of our framework.

For the CoT Tuning (CT) stage, we use our proposed NER-CoT dataset, which contains 45,787 training samples. This dataset is specifically designed to train the model to generate chain-of-thought reasoning for NER tasks.

For the RE stage, we sample from the InstructUIE \cite{wang2023instructuie} collection, which comprises 20 distinct public NER datasets: ACE2005 \cite{doddington2004automatic}, AnatEM \cite{pyysalo2014anatomical}, BC2GM \cite{kocaman2021biomedical}, BC4CHEMD \cite{krallinger2015chemdner}, BC5CDR \cite{li2016biocreative}, Broad Twitter \cite{derczynski2016broad}, CoNLL-2003 \cite{sang2003introduction}, FabNER \cite{kumar2022fabner}, FindVehicle \cite{guan2024findvehicle}, GENIA \cite{kim2003genia}, HarveyNER \cite{chen2022crossroads}, MIT-Movie \cite{liu2021crossner}, MIT-Restaurant \cite{liu2021crossner}, MultiNERD \cite{tedeschi2022multinerd}, NCBI-Disease \cite{dougan2014ncbi}, OntoNotes 5.0 \cite{hovy2006ontonotes}, PolyglotNER \cite{al2015polyglot}, TweetNER7 \cite{ushio2022named}, WikiANN \cite{pan2017cross}, and WikiNeural \cite{tedeschi2021wikineural}.

From this collection, we created a 4,703-sample RE training set using a capped stratified sampling strategy. This method ensures that smaller datasets are adequately represented while preventing exceptionally large datasets (e.g., OntoNotes 5.0, WikiANN) from dominating the RE training mix. The procedure is as follows: (1) For each of the 20 datasets, we determine its ``effective size'' by setting $N_i^{\prime}=\min(N_i,10000)$, where $N_i$ is the original dataset size. (2) We calculate the total effective size $N_{total}^{\prime}=\sum_{i=1}^{20}N_i^{\prime}$. (3) Each dataset's sampling proportion is then calculated as $P_i = N_i^{\prime} / N_{total}^{\prime}$. (4) Finally, we draw $S_i = \lfloor 4703 \times P_i \rfloor$ samples from each dataset $i$ to assemble the final RE dataset.

Our evaluation suite assesses both in-domain (ID) and out-of-domain (OOD) performance. The ID evaluation uses the standard test sets for MIT-Movie and MIT-Restaurant. As these datasets are part of the InstructUIE collection, their training data was included in our RE sampling pool. The OOD evaluation assesses generalization to unseen domains. For this, we use five datasets from the CrossNER benchmark \cite{liu2021crossner}: AI, literature, music, politics, and science. These domains are not represented in the RE training data. Detailed statistics for all datasets are summarized in Table \ref{table1}.

\begin{table}[h]
	\small
	\centering
	\caption{Statistics of the datasets. ``\#RE'' indicates the number of data samples from the RE phase, ``\#SFT'' indicates the number of samples for domain-specific fine-tuning (for Table \ref{table:supervised}), and ``\#Test'' indicates the number of evaluation samples.}
	\label{table1}
	\begin{tabular}{c|cc|ccc}
		\toprule
		\textbf{Dataset} & \textbf{RE} & \textbf{Test} & \textbf{\#RE} & \textbf{\#SFT} & \textbf{\#Test} \\ \midrule
		ACE 2005 & \checkmark & \checkmark & 319 & 7,134 & 1,050 \\
		AnatEM & \checkmark & \checkmark & 253 & 5,667 & 3,758 \\
		BC2GM & \checkmark & \checkmark & 438 & 12,392 & 4,977 \\
		BC4CHEMD & & \checkmark & / & 20,000 & 26,204 \\
		BC5CDR & \checkmark & \checkmark & 199 & 4,545 & 4,788 \\
		Broad Twitter & \checkmark & \checkmark & 233 & 5,324 & 2,000 \\
		CoNLL-2003 & \checkmark & \checkmark & 434 & 12,613 & 3,184 \\
		FabNER & \checkmark & \checkmark & 413 & 9,421 & 2,064 \\
		FindVehicle & \checkmark & \checkmark & 438 & 20,000 & 20,769 \\
		GENIA & \checkmark & \checkmark & 438 & 14,966 & 1,850 \\
		HarveyNER & \checkmark & \checkmark & / & 3,553 & 1,260 \\
		MultiNERD & \checkmark & \checkmark & 438 & 20,000 & 9,994 \\
		NCBI-Disease & \checkmark & \checkmark & 237 & 5,432 & 940 \\
		OntoNotes 5.0 & \checkmark & \checkmark & 425 & 20,000 & 7,782 \\
		PolyglotNER & & \checkmark & / & 20,000 & 10,000 \\
		TweetNER7 & & \checkmark & / & 7,111 & 576 \\
		WikiANN & \checkmark & \checkmark & 438 & 20,000 & 10,000 \\
		WikiNeural & & \checkmark & / & 20,000 & 11,597 \\ \midrule
		CrossNER\_AI & & \checkmark & / & / & 431 \\
		CrossNER\_literature & & \checkmark & / & / & 416 \\
		CrossNER\_music & & \checkmark & / & / & 465 \\
		CrossNER\_politics & & \checkmark & / & / & 650 \\
		CrossNER\_science & & \checkmark & / & / & 543 \\
		MIT-Movie & & \checkmark & / & 9,707 & 2,441 \\
		MIT-Restaurant & & \checkmark & / & 7,658 & 1,520 \\ \bottomrule
	\end{tabular}
\end{table}

\section{Implementation Details}

This section provides detailed information on the construction of our NER-CoT dataset, including the prompts used, and a comprehensive list of all training hyperparameters.

\subsection{NER-CoT Dataset Construction}

The NER-CoT dataset, which contains 45,787 samples, was generated from the Pile-NER corpus \cite{zhou2023universalner}. As illustrated in Figure \ref{fig:framework} in the main paper, our generation process involves three stages: Re-annotation, Validation, and Consistency.

\begin{figure}[t]
	\centering
	\includegraphics[scale=0.75]{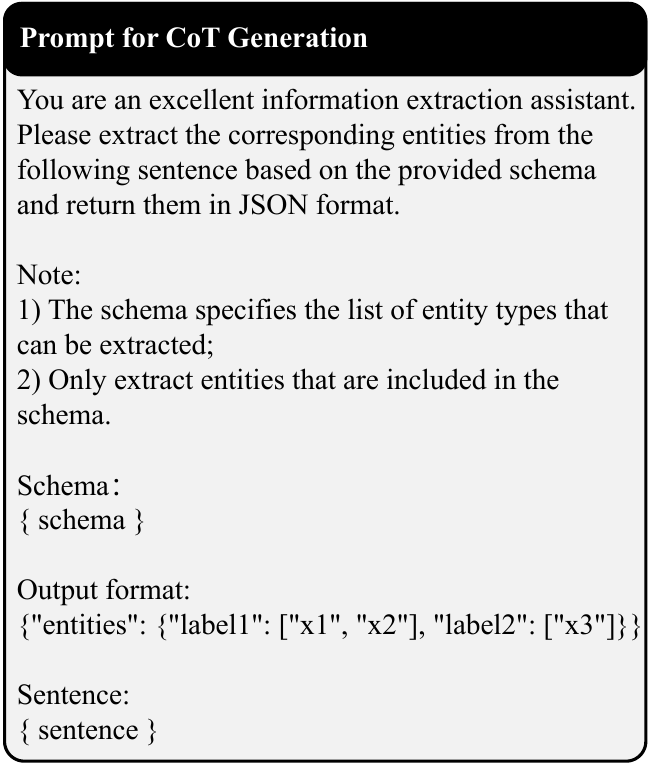}
	\caption{Prompt for CoT Generation.}
	\label{fig:prompt-data}
\end{figure}

\textbf{Re-annotation Step} In this initial stage, we use the sentences and entity schemas from Pile-NER to prompt a large language model (DeepSeek-R1 \cite{guo2025deepseek}) to generate both the extracted entities and an explicit reasoning path. This step produces a raw dataset $\mathcal{D}_{\mathrm{raw}}$. The prompt template used to guide the LLM is shown in Figure \ref{fig:prompt-data}.

\textbf{Validation Step} The raw outputs $\mathcal{D}_{\mathrm{raw}}$ may contain logical fallacies or structural errors. We apply a strict structural validation check. A sample $(X, \mathcal{S}, \mathcal{C}_{\mathrm{raw}}, E_{\mathrm{raw}})$ is retained only if its reasoning trace $\mathcal{C}_{\mathrm{raw}}$ explicitly justifies every entity in the list $E_{\mathrm{raw}}$, and the list $E_{\mathrm{raw}}$ is well-formatted and fully compliant with the predefined entity schema $\mathcal{S}$. Samples failing these checks are discarded.

\begin{figure}[t]
	\centering
	\includegraphics[scale=0.62]{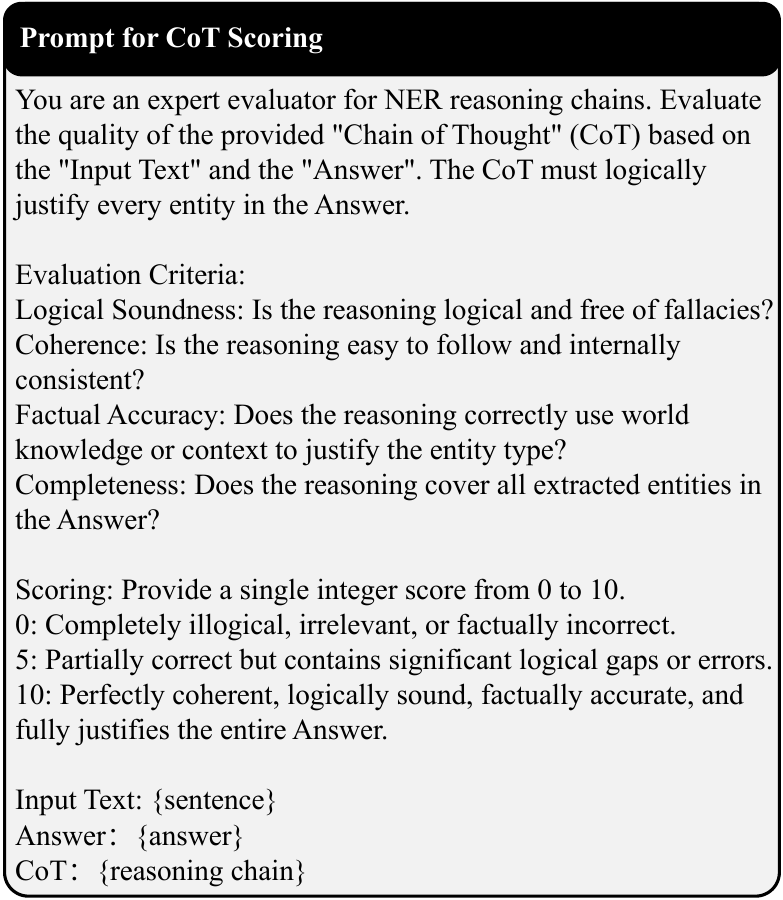}
	\caption{Prompt for CoT Scoring.}
	\label{fig:prompt-score}
\end{figure}

\textbf{Consistency Step} This step evaluates the semantic quality of the validated reasoning paths. We employ an auxiliary LLM (Qwen3 32B) to assess the internal coherence, logical soundness, and factual accuracy of the reasoning. This model is prompted to assign a consistency score from 0 to 10. The prompt for this evaluation is shown in Figure \ref{fig:prompt-score}.

We set a high threshold, retaining only samples with a consistency score of 9 or higher. This rigorous three-step process yields the final NER-CoT dataset, $\mathcal{D}_{\mathrm{cot}}$, which is used for the CoT Tuning stage.

\subsection{Training and Hyperparameter Details}

We use the qwen3-8B-Base \cite{qwen3} model as the foundation for all experiments, loaded via the transformers library \cite{wolf2020transformers}. To ensure reproducibility, the global random seed is set to 42. We utilize bfloat16 mixed-precision training, gradient checkpointing, FlashAttention-2 \cite{dao2023flashattention}, and the Liger-kernel \cite{hsu2024liger} for computational efficiency. The AdamW optimizer is employed across all training stages.

\textbf{Stage 1: CoT Tuning (CT)} This stage is conducted for five epochs. The maximum sequence length is set to 8192 tokens. We use a learning rate of $2 \times 10^{-5}$ with a cosine learning rate scheduler and a warmup ratio of 0.03. The weight decay is set to 0.0, and the gradient clipping norm is 1.0. Training is performed on 8 A800 GPUs, with a per-GPU batch size of 32, resulting in an effective global batch size of 256.

\textbf{Stage 2: Reasoning Enhancement (RE)} This stage is optimized using the GRPO reinforcement learning algorithm for one epoch. The GRPO clipping threshold $\epsilon$ is set to 0.2, and the KL coefficient $\beta$ is set to 0.04. Training is conducted on 6 A800 GPUs, with a per-GPU batch size of 8 and 8 gradient accumulation steps, resulting in an accumulated total batch size of 384 ($6 \times 8 \times 8$). An additional 2 A800 GPUs are allocated for the rollout process. During rollout, the temperature is set to 0.9, and 16 candidate outputs are generated for each prompt, with the maximum output length constrained to 4096 tokens. The reward function weights are set as $\lambda_{\text{F1}}=10$ and $\lambda_{\text{schema}}=1$.

\section{More Experiments}

\subsubsection{Cross-Lingual Generalization Analysis}

To further evaluate the generalization capabilities of our proposed model, we conducted a comprehensive cross-lingual Named Entity Recognition evaluation on a dataset spanning 11 distinct languages. The results are presented in Table 6-4. The languages evaluated include English (en), Chinese (zh), German (de), Spanish (es), Dutch (nl), Russian (ru), Bengali (bn), Farsi (fa), Hindi (hi), Korean (ko), and Turkish (tr). The evaluation metrics include language-specific F1 scores, the average cross-lingual performance ($Avg_{cross}$), and the overall average performance ($Avg$).

\begin{table*}[t]
	\centering
	\caption{Results of cross-lingual evaluation in Multiconer22. $Avg_{cross}$ measures the average zero-shot transfer performance on the nine unseen languages. $Avg$ denotes the overall average performance across all 11 languages.}
	\label{table:lingual}
	\begin{tabular}{l|cc|ccccccccc|cc}
		\toprule
		\multirow{2}{*}{\textbf{Model}}        &         &         & \multicolumn{9}{c|}{\textbf{Cross-Lingual}}                                                                                                                                  & \multirow{2}{*}{\textbf{$\text{Avg}_{cross}$}} & \multirow{2}{*}{\textbf{Avg}} \\ \cmidrule{4-12}
		& \textbf{en} & \textbf{zh} & \textbf{de} & \textbf{es} & \textbf{nl} & \textbf{ru} & \textbf{bn} & \textbf{fa} & \textbf{hi} & \textbf{ko} & \textbf{tr} &                                     &                               \\ \midrule
		ChatGPT                             & 37.2             & 18.8             & 37.1            & 34.7             & 35.7           & 27.4             & 23.3             & 25.9             & 27.3           & 30.0            & 31.9             & 30.4                                & 29.9                          \\
		YAYI-UIE                            & 52.0             & 37.9             & 41.8            & 42.4             & 38.0           & 31.0             & 16.0             & 22.0             & 20.9           & 26.5            & 30.7             & 29.9                                & 32.7                          \\
		IEPILE                              & 53.2             & 39.3             & 39.9            & 42.4             & 35.7           & 28.7             & 12.3             & 23.9             & 19.0           & 27.6            & 26.9             & 28.5                                & 31.7                          \\
		GLiNER                              & 41.7             & 24.3             & 39.5            & 42.1             & 38.9           & 33.3             & 25.9             & 30.2             & 27.8           & 28.7            & 30.0             & 32.9                                & 32.9                          \\
		$\text{B}^2\text{NER-7B}$ & 54.8             & 45.4             & 36.6            & 46.0             & 43.0           & 33.9             & -                & -                & -              & -               & -                & -                                   & -                             \\
		KnowCoder-X                         & 56.4             & 47.5             & 49.9            & 54.2             & 48.8           & 41.5             & 24.1             & 30.8             & 29.7           & 33.6            & 43.2             & 39.5                                & 41.8                          \\ \midrule
		ReasoningNER 8B                      & \textbf{59.7}    & \textbf{53.4}    & \textbf{58.4}   & \textbf{56.0}    & \textbf{53.2}  & \textbf{47.5}    & \textbf{40.7}    & \textbf{38.6}    & \textbf{46.2}  & \textbf{44.4}   & \textbf{47.6}    & \textbf{48.1}                       & \textbf{49.6} \\ \bottomrule
	\end{tabular}
\end{table*}

As shown in Table \ref{table:lingual}, ReasoningNER-8B achieves an average F1 score of 48.1\% on the nine unseen languages, significantly outperforming all comparative models. This represents a substantial improvement of 8.6 percentage points over the previous state-of-the-art model, KnowCoder-X. It is crucial to note that ReasoningNER-8B was trained exclusively on English (en) data, whereas all baseline methods utilized bilingual (en+zh) training data. Despite this disadvantage, ReasoningNER-8B's performance on the Chinese (zh) test set still surpasses all baselines that were explicitly trained on Chinese data.These findings strongly suggest that the reasoning paradigm proposed in this paper guides the model to acquire a more fundamental and language-agnostic extraction capability. Instead of relying on superficial, language-specific lexical or syntactic features, the model learns to understand entity definitions and contextual relationships through explicit logical inference. This mechanism facilitates the effective and robust transfer of extraction knowledge to novel, unseen languages.

\section{Limitions}

\begin{figure*}[t]
	\centering
	\includegraphics[scale=0.7]{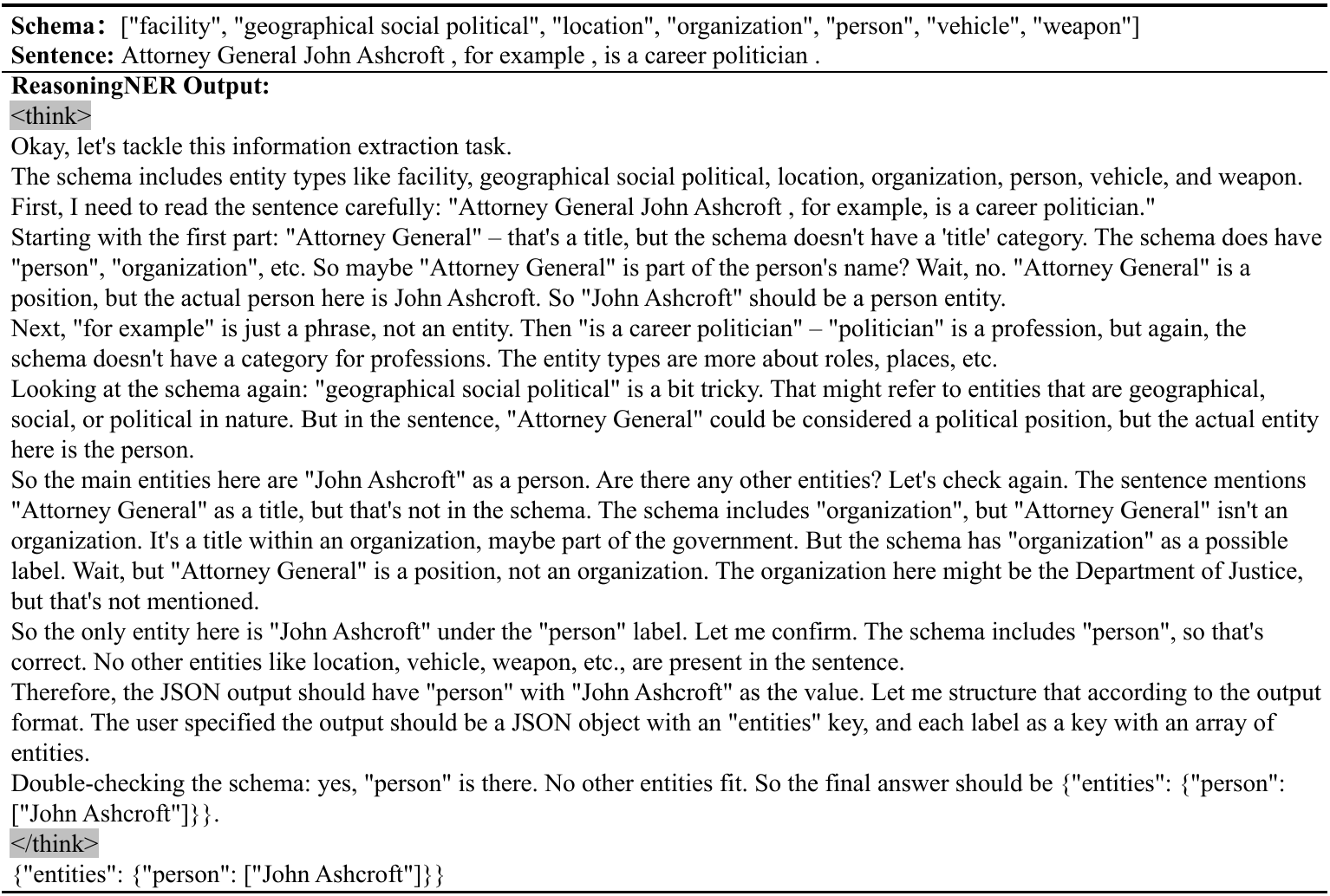}
	\caption{Example of ReasoningNER's Chain-of-Thought Output.}
	\label{fig:output-example}
\end{figure*}

Despite the compelling performance of ReasoningNER in zero-shot and few-shot NER, the current work possesses limitations that delineate clear avenues for future research. A primary observation is the verbosity and redundancy inherent in the generated CoT processes. As illustrated by Figure \ref{fig:output-example}, the model often produces extensive, multi-step rationales encompassing detailed analysis, exclusion criteria, and self-correction, even for relatively short sentences with a limited number of target entity types. While this exhaustive reasoning contributes to accuracy, it incurs significant computational overhead and excessive token consumption, thereby impeding scalability for long-document processing and large-scale datasets. Consequently, a critical direction for future work is the development of CoT compression techniques. This research will explore methodologies to reduce token expenditure while preserving inferential integrity, potentially through advanced prompt engineering to elicit more succinct reasoning paths or post-processing algorithms designed to prune redundant steps from the generated rationales.

Furthermore, the current framework's capabilities are principally concentrated on Named Entity Recognition. Information Extraction is a broader field encompassing more complex tasks, such as relation extraction and event extraction, which our current model does not address. We thus plan to extend this reasoning-based paradigm into a unified framework for Universal Information Extraction. The objective is to construct a single, versatile model capable of not only identifying entities but also inferring the semantic relations between them and extracting structured event information. This extension will necessitate the design of more sophisticated reasoning patterns to capture the intricate dependencies between entities, relations, and events, thereby further unlocking the potential of large language models for complex, reasoning-oriented structured information extraction.

\end{document}